\documentclass[10pt,twocolumn,letterpaper]{article}

\usepackage[pagenumbers]{iccv} % To force page numbers, e.g. for an arXiv version
\usepackage{pifont}
\usepackage{multirow}
\usepackage[normalem]{ulem}
\useunder{\uline}{\ul}{}
\definecolor{iccvblue}{rgb}{0.21,0.49,0.74}
\usepackage[pagebackref,breaklinks,colorlinks,allcolors=iccvblue]{hyperref}

\def\paperID{8787} % *** Enter the Paper ID here
\def\confName{ICCV}
\def\confYear{2025}

\title{DeepDubber-V1: Towards High Quality and Dialogue, Narration, Monologue Adaptive Movie Dubbing Via Multi-Modal Chain-of-Thoughts Reasoning Guidance.}

\author{
    \textit{Junjie Zheng$^1$$^*$, Zihao Chen$^{1}$$^*$, Chaofan Ding$^1$, Xinhan Di$^1$} \\
    $^1$AI Lab, Giant Network.\\
    $^2$University of Trento.\\
    \small\texttt{\{zhengjunjie, chenzihao, dingchaofan, dixinhan\}@ztgame.com}
}

\begin{document}
\maketitle
\begin{abstract}
% The ABSTRACT is to be in fully justified italicized text, at the top of the left-hand column, below the author and affiliation information.
% Use the word ``Abstract'' as the title, in 12-point Times, boldface type, centered relative to the column, initially capitalized.
% The abstract is to be in 10-point, single-spaced type.
% Leave two blank lines after the Abstract, then begin the main text.
% Look at previous \confName abstracts to get a feel for style and length.

Current movie dubbing technology can generate the desired voice from a given speech prompt, ensuring good synchronization between speech and visuals while accurately conveying the intended emotions. However, in movie dubbing, key aspects such as adapting to different dubbing styles, handling dialogue, narration, and monologue effectively, and understanding subtle details like the age and gender of speakers, have not been well studied. To address this challenge, we propose a framework of multimodal large language model. First, it utilizes multimodal Chain-of-Thought (CoT) reasoning methods on visual inputs to understand dubbing styles and fine-grained attributes. Second, it generates high-quality dubbing through large speech generation models, guided by multimodal conditions. Additionally, we have developed a movie dubbing dataset with CoT annotations. The evaluation results demonstrate a performance improvement over state-of-the-art methods across multiple datasets. In particular, for the evaluation metrics, the SPK-SIM and EMO-SIM increases from $82.48\%$ to $89.74\%$, $66.24\%$ to $78.88\%$ for dubbing setting 2.0 on V2C-Animation dataset, LSE-D and MCD-SL decreases from 14.79 to 14.63, 5.24 to 4.74 for dubbing setting 2.0 on Grid dataset, SPK-SIM increases from 64.03 to 83.42 and WER decreases from 52.69\% to 23.20\% for initial reasoning setting on proposed CoT-Movie-Dubbing dataset in the comparison with the state-of-the art models.

\end{abstract}    
\section{Introduction}

Dubbing involves adding the correct human voice to a video's dialogue, ensuring synchronization with the character's lip movements, and conveying the emotions of the scene. It plays a vital role in film, television, animation and gaming, enhancing immersion and effectively conveying emotions and atmosphere. Existing dubbing methods can be categorized into two groups, both of which focus on learning different styles of key prior information to generate high-quality voices. The first group focuses on learning effective speaker style representations \cite{chen2022v2c,hassid2022more,wan2018generalized,cong2023learning}. The second group aims to learn appropriate prosody by utilizing visual information from the given video input \cite{cong2023learning,hu2021neural,lee2023imaginary,zhao2024mcdubber}. However, the accuracy of these priors is insufficient and inadequate for movie dubbing in real-world scenarios. For example, adaptive dubbing for different types, such as dialogue, narration, and monologue, as well as fine-grained attributes such as expected ages and genders, has not been thoroughly studied \cite{cong2024styledubbermultiscalestylelearning,hu2021neural}.

%due to the limitations of various modules, which impacts the quality of speech generation. 
%Moreover, the currently studied priors are inadequate for movie dubbing in real-world scenarios. For example, adaptive dubbing for different types, such as dialogue, narration, and monologue, as well as fine-grained attributes such as expected ages and genders, has not been thoroughly studied \cite{cong2024styledubbermultiscalestylelearning,hu2021neural}.

\begin{figure}[t!]
    \centering
    \resizebox{\linewidth}{!}{
    \includegraphics[]{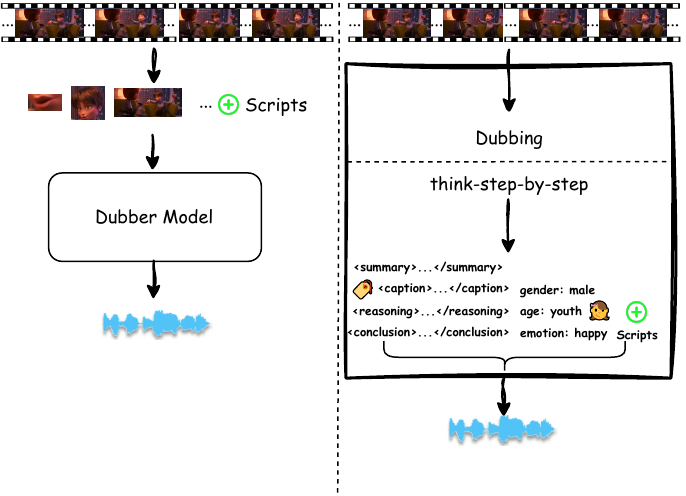}
    }
    \caption{Curent Dubbing models \cite{congLearningDubMovies2023,cong2024styledubbermultiscalestylelearning,zhang2024from} (Left). Proposed Dubbing Models (Right) For dubbing types and fine-grained attributes.}
    \label{fig:model_pipeline}
\end{figure}

With the rapid advancement of large language reasoning models with step-by-step thinking ability \cite{o1-min,o1,GPT-o3-mini,grok-3,claude-3-7,deepseekai2025deepseekr1incentivizingreasoningcapability,QwQ-32B}
% (+o1-mini,o1,o3,o3-mini,grok3,cloud3.7,kimi1.6,R1,QWQ-32B-RL) 
and methods that enhance reasoning capabilities to interpret visual information through CoT, MLLM have increasingly shown their potential in multimodal reasoning and understanding tasks \cite{penamakuri2024AudiopediaAudioQA,Insight,cheng2024videollama2advancingspatialtemporal,kim2024videoiclconfidencebasediterativeincontext,bonomo2025visualragexpandingmllm,liu2024EnhancingVisualReasoning,li2025imaginereasoningspacemultimodal,sahili2024FairCoTEnhancingFairness,zheng2024ThinkingLookingImproving}. These advancements in reasoning capabilities within MLLM hold promise for accurately providing dubbing types and fine-grained attributes.

Therefore, we propose a multimodal large language model for high-quality movie dubbing that effectively understands dubbing styles and fine-grained attributes. First, through multi-modal CoT learning, a multimodal large language model is trained to improve its reasoning ability, enabling a better understanding of dubbing types (dialogue, narration, monologue) and fine-grained attributes from video inputs. Secondly, a large multimodal speech generation model is trained with designed control mechanisms using multiple-modal conditions. Thirdly, we create a CoT multi-modal movie dubbing dataset annotated with step-by-step reasoning instructions.

\section{Related Work}

\subsection{Visual Voice Cloning}
Current advanced dubbing technologies significantly enhance speech-video synchronization and emotional expression by integrating visual and textual information.
Some works focus on improving speaker identity to handle multi-speaker scenes \cite{cong2024emodubberhighqualityemotion,congLearningDubMovies2023,zhang2024from,cong2024styledubbermultiscalestylelearning}. For example, Speaker2Dub \cite{zhang2024from} introduces speaker embedding extracted by pre-trained GE2E to the phoneme encoder and the mel spectrogram decoder by a learnable style affine transform, while StyleDubber \cite{cong2024styledubbermultiscalestylelearning} proposes a multi-scale style adapter with phoneme and utterance level to strengthen speaker characteristics. In addition, some works attempt to combine visual representation to enhance prosody expressive \cite{congLearningDubMovies2023,hu2021neural,lee2023imaginary,zhao2024mcdubber}. For example, HPMDubbing \cite{congLearningDubMovies2023} is a hierarchical dubbing method that bridges acoustic details with visual information: lip motion, face region, and scene. To improve contextual prosody, MCDubber \cite{zhao2024mcdubber} enlarges the modeling object from a single sentence to the previous and following sentences, incorporating more contextual video scenes. Although speaker identity and prosody modeling have received attention, existing works still suffer from poor lip-sync and lifeless emotional expression, which is unacceptable in dubbing.

\subsection{Flow-Matching Speech Generation}
Flow Matching \cite{lipman2023flow} is a simulation-free method to train Continuous Normalizing Flows (CNFs) \cite{ChenRBD18} models, which model arbitrary probability path and capture the probability trajectories represented by diffusion processes \cite{song2021maximum}. Due to its advantages of high sampling speed and generation quality, flow matching has attracted significant attention in speech generation \cite{10445948,10448291,kim2023pflow}. Recently, Matcha-TTS \cite{10448291} and DiTTo-TTS \cite{lee2025dittotts} have introduced optimal-transport conditional flow matching (OT-CFM) for training, which yields an ODE-based decoder to improve the fidelity of the mel spectrograms. Then, F5TTS \cite{chen2024f5ttsfairytalerfakesfluent} leverages the Diffusion Transformer with ConvNeXt V2 \cite{10205236} to better tackle text-speech alignment during in-context learning. However, these works are limited in the field of TTS and cannot be applied to the V2C task. Therefore, we study the integration with MLLM reasoning models and TTS for the V2C task.

\subsection{Chain-of-thought Reasoning}
Visual reasoning demands the model’s visual perception capability and high-level cognition ability \cite{Johnson2016CLEVRAD,MALKINSKI2023713}. Several tasks have been applied to evaluate the visual reasoning ability of Visual-Language Models (VLMs), including VQA \cite{90339209302102270,9859766} requiring models to answer visual content and textual questions, and Visual Entailment \cite{Song2022CLIPMA,323233524965,33333978-3-031-73229-4_13} requiring models to determine the consistency of text descriptions and visual content, etc. With the development of LLMs, vision-language models leverage the advanced reasoning abilities of LLMs to interpret visual tasks \cite{liu2023visual,ccc2405-20224}. Some vision-language models enhance visual reasoning by optimizing the visual encoding strategy \cite{10.1016/j.inffus.2024.102270,zhou2025lvp,liu2023visual,Gupta2022VisualPC,Hu2024FinetuningLL,10.1145/3544548.3581388} to produce cognition-focused visual tokens. Then, with the rapid advancement of large language reasoning models  with the step-by-step thinking ability \cite{o1-min,o1,GPT-o3-mini,grok-3,claude-3-7,deepseekai2025deepseekr1incentivizingreasoningcapability,QwQ-32B}, vision-language task is studied through step-step reasoning for a variety of multimodal large language models. However, step-step reasoning mechanism is not well-studied in the movie dubbing, therefore, we propose DeepDubbber for movie dubbing with internal multimodal chain-of-thoughts reasoning guidance.

\begin{figure*}[h]
    \centering
    \includegraphics[scale=0.7]{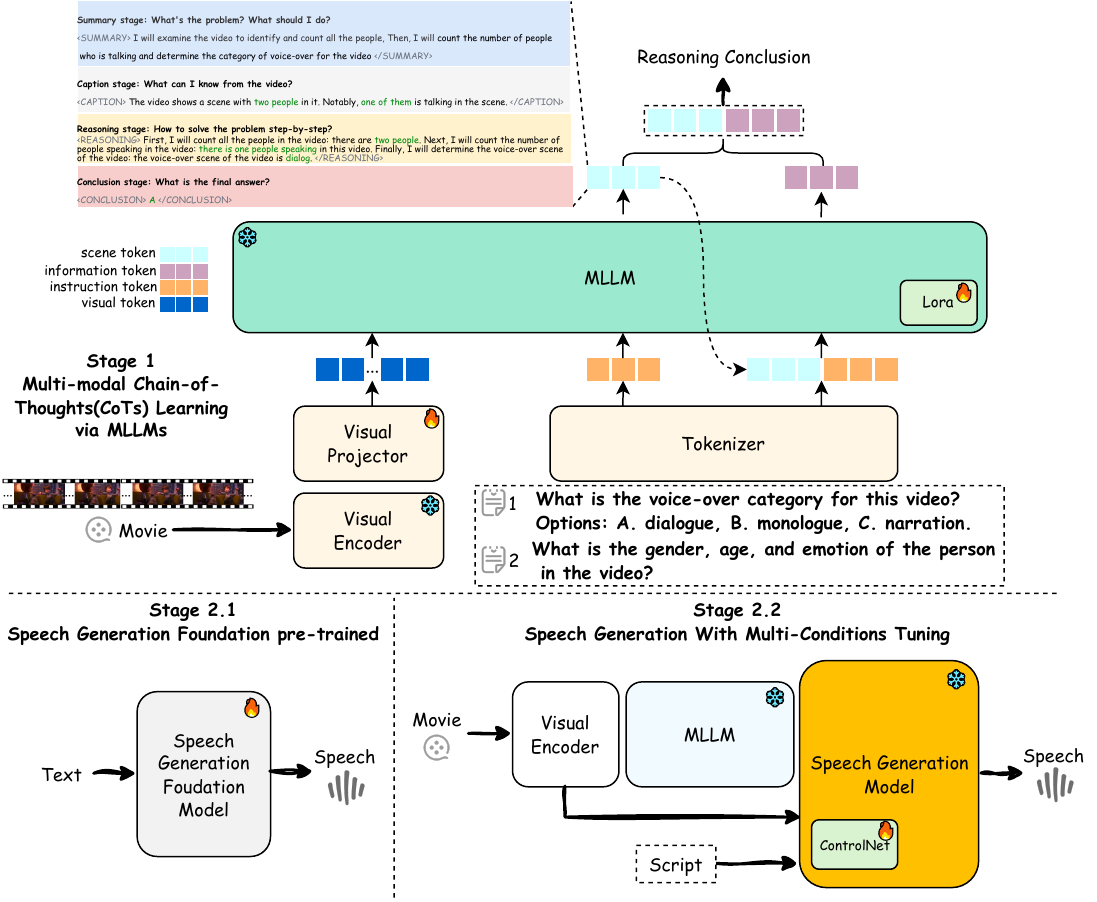}
    \caption{DeepDubber pipeline with multi-stage, multi-modal training.}
    \label{fig:training_pipeline}
\end{figure*}

\section{Method}

\subsection{Overview}

Given a silent video clip $V_l$, a corresponding subtitle $T_v$, and the goal of generating a fully dubbed video, the proposed model (DeepDubber) aims to produce speech $\hat{S}$ that matches the video, ensures contextual and prosodic relevance, and maintains speech-video synchronization with the help of MLLM. The model can be formalized as follows:
\begin{equation}
\hat{S} = F_{dubber}(V_l, T_v)
\end{equation} 
DeepDubber consists of two modeling stages: i) Multimodal reasoning and understanding through in-context learning and mixed preference optimization. 
ii) Speech Generation Stage: This stage incorporates a conditional DiT based speech generator. 

\subsection{Multi-modal Chain-of-Thought Learning via MLLM}

\subsubsection{Stage 1.1: Training Multi-modal Chain-of-Thought via Supervised Learning.}  
The core functionality of DeepDubber is to extract key semantic features from the visual stream that are crucial for the dubbing process. These features include scene type, speaker gender, speaker age, and speaker emotion, which are inferred step by step, as illustrated in Figure \ref{fig:dataset_cot}. Inspired by the success of MLLM, such as VLMs~\cite{liu2023visual,xu2025llavacotletvisionlanguage,Wang2024EnhancingTR}, we leverage multimodal instruction tuning to enhance CoT reasoning, thereby improving the quality of movie dubbing. The CoT reasoning process is formulated as follows:  
\small
\begin{flalign}
    C_1^{CoT} &= F_{mcot}^1(V, \text{Instruct}_1), \\
    C_2^{QA} &= F_{qa}^2(V, \text{Instruct}_{2}) 
\end{flalign}
\normalsize
\noindent where \( V \) represents the input video clip, \( \text{Instruct}_i \) denotes the \( i \)-th instruction provided to guide the reasoning process, \( F_{\text{mcot}}^i \) is the function representing the \( i \)-th step of multimodal CoT reasoning using an MLLM, \( C_1^{CoT} \) and \( C_2^{QA} \) are intermediate outputs from the reasoning process, where \( C_1^{CoT} \) is the result of the first CoT reasoning step, and \( C_2^{QA} \) is obtained through a question-answering (QA) step.

To optimize the model’s response generation, we define the multimodal reasoning process as:
% \resizebox{\linewidth}{!}{
\begin{flalign}
    & \text{M}_{\text{mllm}} : (\text{Video}_{\text{clip}}, \text{CoT}_{\text{instruction}}, \text{QA}_{\text{instruction}}) \mapsto \text{Response}, & \\
    & \min_{\theta_{\text{response}}} \mathbb{E}_{(\text{Video}_{\text{clip}}, \text{CoT}_{\text{instruction}}, \text{QA}_{\text{instruction}}) \sim \mathcal{D}}
    \Big[ \mathcal{L}_{res} \Big( \text{M}_{\text{mllm}}(\text{Video}_{\text{clip}}, & \notag \\
    & \quad \text{CoT}_{\text{instruction}}, \text{QA}_{\text{instruction}}), \text{Response}_{\text{gt}} \Big) \Big], &
\end{flalign}
% }
\noindent where \( \text{M}_{\text{mllm}} \) represents the multimodal large language model performing CoT reasoning and QA, \( \text{Video}_{\text{clip}} \) is the input video segment, \( \text{CoT}_{\text{instruction}} \) and \( \text{QA}_{\text{instruction}} \) are instructions guiding the CoT reasoning and QA process, respectively, \( \text{Response} \) is the generated output of the model, \( \mathcal{D} \) denotes the distribution of the training dataset, \( \theta_{\text{response}} \) represents the parameters of the model to optimize, \( \mathcal{L}_{res} \) is the loss function measuring the difference between the predicted response of the model and the ground truth response \( \text{Response}_{\text{gt}} \).
This approach ensures that the MLLM effectively reasons over multimodal inputs, facilitating high-quality dubbing through structured stepwise reasoning.

\subsubsection{Stage 1.2: Training Multi-modal Chain-of-Thought via Reinforcement Learning.}
%follow R1 RL Scaling + Performance Learning.
The reward is the source of the training signal that decides the direction of RL optimization. To train CoT-MLLM, we adopt a rule-based reward system like Deepseek-R1 \cite{deepseekai2025deepseekr1incentivizingreasoningcapability} that consists mainly of two types of rewards: accuracy rewards and format rewards. We employ a format reward model that enforces the model to put its reasoning process between \texttt{<SUMMARY>} \texttt{</SUMMARY>}, \texttt{<CAPTION>}\texttt{</CAPTION>}, \texttt{<REASONING>}\texttt{</REASONING>}, and \texttt{<CONCLUSION>} \texttt{</CONCLUSION>} tags.
We use the Mixed Preference Optimization (MPO) \cite{Wang2024EnhancingTR} method to learn the relative preferences between pairs of responses and enhance the reasoning capability of MLLM across different instructions. The mixed preference optimization is applied to further enhance the ability of the multimodal CoT reasoning. The training objective is represented as the following.

\noindent\textbf{Training Objective.}
% The Mixed Preference Optimization (MPO) objective combines three loss components and F\&O rewards: preference loss $L_p$, quality loss $L_q$, generation loss $L_g$, format loss $L_f$ and accuracy loss $L_c$. 
The MPO objective combines three loss components and F\&O rewards: preference loss (\(L_p\)), quality loss (\(L_q\)), generation loss (\(L_g\)), format loss (\(L_f\)), and accuracy loss (\(L_c\)).
The total loss is formulated as:
\begin{equation}
    L = w_p L_p + w_q L_q + w_g L_g + w_fL_f + w_cL_c,
\end{equation}
where $w_*$ represents the weight for each loss. We use DPO \cite{rafailov2023direct} for preference loss and BCO \cite{Chen2024LowRedundantOF} for quality loss.
The details of three terms of the loss are then represented as the following:

\noindent\textbf{Preference Loss.}
The DPO \cite{rafailov2023direct} loss models the relative preference between the chosen and rejected responses without requiring a reward model. The loss function is:
\begin{equation}
    L_p = -\log \sigma \left( \beta \log \frac{\pi_\theta(y_c \mid x)}{\pi_0(y_c \mid x)} - \beta \log \frac{\pi_\theta(y_r \mid x)}{\pi_0(y_r \mid x)} \right),
\end{equation}
where $\beta$ is the KL penalty coefficient, $x$ is the user query, $y_c$ is the chosen response, $y_r$ is the rejected response, and $\pi_\theta$ is the policy model initialized from $\pi_0$.

\noindent\textbf{Quality Loss.}
The BCO \cite{Chen2024LowRedundantOF} loss measures the absolute quality of individual responses using a binary classifier. The total loss is:
\begin{equation}
    L_q = L_q^+ + L_q^-,
\end{equation}
where the chosen and rejected loss terms are:
\begin{equation}
    L_q^+ = -\log \sigma \left( \beta \log \frac{\pi_\theta(y_c \mid x)}{\pi_0(y_c \mid x)} - \delta \right),
\end{equation}
\begin{equation}
    L_q^- = -\log \sigma \left( -\left( \beta \log \frac{\pi_\theta(y_r \mid x)}{\pi_0(y_r \mid x)} - \delta \right) \right),
\end{equation}
and $\delta$ is the reward shift for stabilizing training.

\noindent\textbf{Generation Loss.}
The SFT loss helps the model learn to generate preferred responses. The loss is defined as:
\begin{equation}
    L_g = -\frac{\log \pi_\theta(y_c \mid x)}{|y_c|}.
\end{equation}

\noindent\textbf{Format Reward.}
The Format loss helps the model to learn to generate in preferred format. The loss is defined as:
\begin{equation}
    L_f = -\sum \left[ f_{\text{true}} \log(p_f + (1 - f_{\text{true}}) \log(1 - p_f) \right]
\end{equation}
$f_\text{true} \in \{0, 1\}$: format correct or not, $p_f$: the probability of the format being correct predicted by the model.

\noindent\textbf{Outcome Reward.}
The Accuracy loss helps the model learn to generate preferred answer. The loss is defined as:
\begin{equation}
    % \resizebox{\linewidth}{!}{$
    L_o = -\sum \left[ o_{\text{true}} \log(p_o) + (1 - o_{\text{true}}) \log(1 - p_o) \right]
    % $}
\end{equation}
$o_{\text{true}} \in \{0, 1\}$: the answer is correct or not, $p_o$: the probability of the format being correct predicted by the model.

\subsection{Mutli-Conditioned Speech Generation}

\subsubsection{Speech Generation Foundation Pre-Training}
In the second stage of DeepDubber, we first train the foundational speech generation model. To optimize this process, we aim to learn the parameters \(\theta_{\text{generation}}\) by minimizing the composite Conditional Flow Matching (CFM) loss \(\mathcal{L}_{\text{cfm}}\). The speech generation process is formulated as:  
\begin{equation}
\resizebox{\linewidth}{!}{$
\begin{aligned}
    \text{M}_{\text{speech}}: & (\text{Video}_{\text{clip}}, \text{Speech}_{\text{prompt}}, \text{Caption}_{\text{condition}}, \text{Transcript}_{\text{text}}) \\ 
    & \mapsto \text{Speech}_{\text{target}} \quad,
\end{aligned}
$}
\end{equation}
\begin{equation}
\begin{aligned}
    \min_{\theta_\text{generation}} & \mathbb{E}_{(\text{Video}_\text{clip}, \text{Speech}_\text{prompt}, \text{Caption}_\text{condition}, \text{Transcript}_\text{text}) \sim \mathcal{D}} \\
    & \Big[ \mathcal{L}_{cfm} \Big( \text{M}_{\text{speech}}(\text{Video}_{\text{clip}}, \text{Speech}_{\text{prompt}}, \\
    & \quad \quad \quad \text{Caption}_{\text{condition}}, \text{Transcript}_{\text{text}}) \Big) \Big].
\end{aligned}
\end{equation}
We adopt the same architecture as F5-TTS \cite{chen2024f5ttsfairytalerfakesfluent}, which employs a diffusion transformer (DiT) as the backbone. The model is trained to output a vector field \( v_\tau \) using the CFM objective \(\mathcal{L}_{\text{cfm}}\) \cite{lipman2023flow}, defined as:
\begin{equation}
% \resizebox{\linewidth}{!}{$
\mathcal{L}_{\text{cfm}}(\theta) = \mathbb{E}_{\tau, q(x_1), p(x|x_1)} \lVert u_\tau(x|x_1) - v_\tau(x; \theta) \rVert^2
% $}
\end{equation}
\noindent where \( p_\tau \) represents the probability path at time \( \tau \), \( u_\tau \) is the designated vector field for \( p_\tau \), \( x_1 \) is a random variable corresponding to the training data, \( q(x_1) \) denotes the distribution of the training data. By optimizing \(\mathcal{L}_{\text{cfm}}\), the model learns to generate high-quality speech synchronized with the visual and textual cues, ensuring natural and contextually appropriate dubbing.

\subsubsection{Speech Generation With Multi-Conditions Tuning}

Next, in the ControlNet-transformer tuning stage, the video frames, along with an instruction are fed as inputs to the MLLM model. The sequence of video features and the video understanding conclusion are then combined and passed into the speech generation model. In this context, the provided conclusion helps guide the V2S generation process, as shown in stage 2.2 of Figure \ref{fig:training_pipeline}. The proposed speech generation model takes as input the script, silent video, video understanding conclusion, and an optional reference speech, and generates video-aligned speech context sequences, which can be described as:
\begin{equation}
\hat{S} = F_{generator}(V_l, C_v \{C_1, \ldots, C_n\}_{n=4}, T_v)
\end{equation} 
\noindent where $C_v$ is the combination of \(\{C_1, \ldots, C_n\}, n=4\), which represents the scene type condition $C_s$, speaker gender condition $C_g$, speaker age condition $C_a$, and speaker emotion condition $C_e$. These conditions are combined and encoded by an encoder \cite{RaffelSRLNMZLL20}. The $V_l$ represents the visual features derived from the input video frames, which are encoded by CLIP \cite{RadfordKHRGASAM21}.
We implement a cross-attention mechanism that facilitates the integration of understanding conclusion features $C_v$ and visual features $V_l$. 
Furthermore, \(T_v\) represents the embedded script. Furthermore, we added a duration loss \(\mathcal{L}_{dur}\) to constrain the duration consistency, which can be described as:
\begin{equation}
\mathcal{L}_{\text{dur}} = \ell\big( f(V_l, C_l), \, dur \big) ,
\end{equation} 
The final loss function is constructed as follows:
\begin{equation}
    \resizebox{\linewidth}{!}{$
    \mathcal{L}_{\text{g}} = \mathbb{E}_{\tau, q(x_1), p(x|x_1)} \lVert u_\tau(x|x_1, v_l, c_v, t_v) - v_\tau(x; \theta) \rVert^2 + \mathcal{L}_{\text{dur}}
$}
\end{equation}
In the training stage, the visual condition $V_l$, video understanding conclusion condition $C_v$, and video script condition $T_v$ are each set to $\phi$ with a 5 \% probability.
Extending classifier-free guidance from the script condition to visual input and visual understanding enhances both conditional control precision and speech quality. The guidance scales $\lambda_V$, $\lambda_C$, and $\lambda_T$, correspond to the video clip, video conclusion, and video-related script, respectively, and measure the alignment between the sampling results and conditions. Inspired by \cite{jiang2024dive}, during inference, the modified velocity estimate is as follows:
\begin{equation}
\resizebox{0.8\linewidth}{!}{$
\begin{aligned}
g_{\theta}' & = g_{\theta}(x_\tau, \phi, \phi, \phi) \\ 
 & + \lambda_V \cdot (v_0(x_\tau, c_v, c_c, c_t) - v_0(x_\tau, \phi, c_c, c_t)) \\ 
 & + \lambda_C \cdot (v_0(x_\tau, \phi, c_c, c_t) - v_0(x_\tau, \phi, \phi, c_t)) \\ 
 & + \lambda_T \cdot (v_0(x_\tau, \phi, \phi, c_t) - v_0(x_\tau, \phi, \phi, \phi)) \\
\end{aligned}
$}
\end{equation}
\begin{figure}[t!]
    \centering
    \resizebox{\linewidth}{!}{
    \includegraphics[]{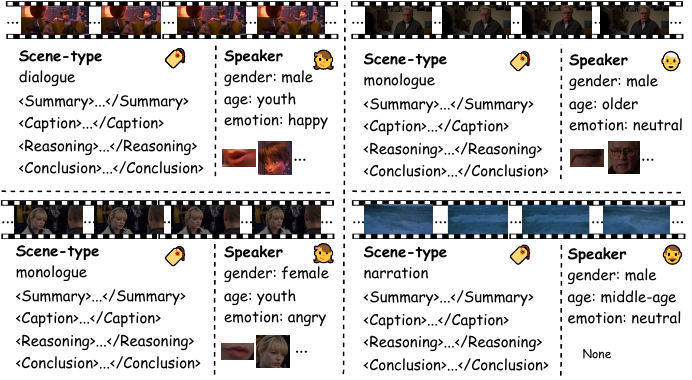}
    }
    \caption{Proposed dataset with multi-type annotations, including annotation for lips, faces, scene-type, speaker gender, speaker age, voice emotion.}
    \label{fig:dataset}
\end{figure}

\begin{figure*}[t!]
    \centering
    \resizebox{\linewidth}{!}{
    \includegraphics[]{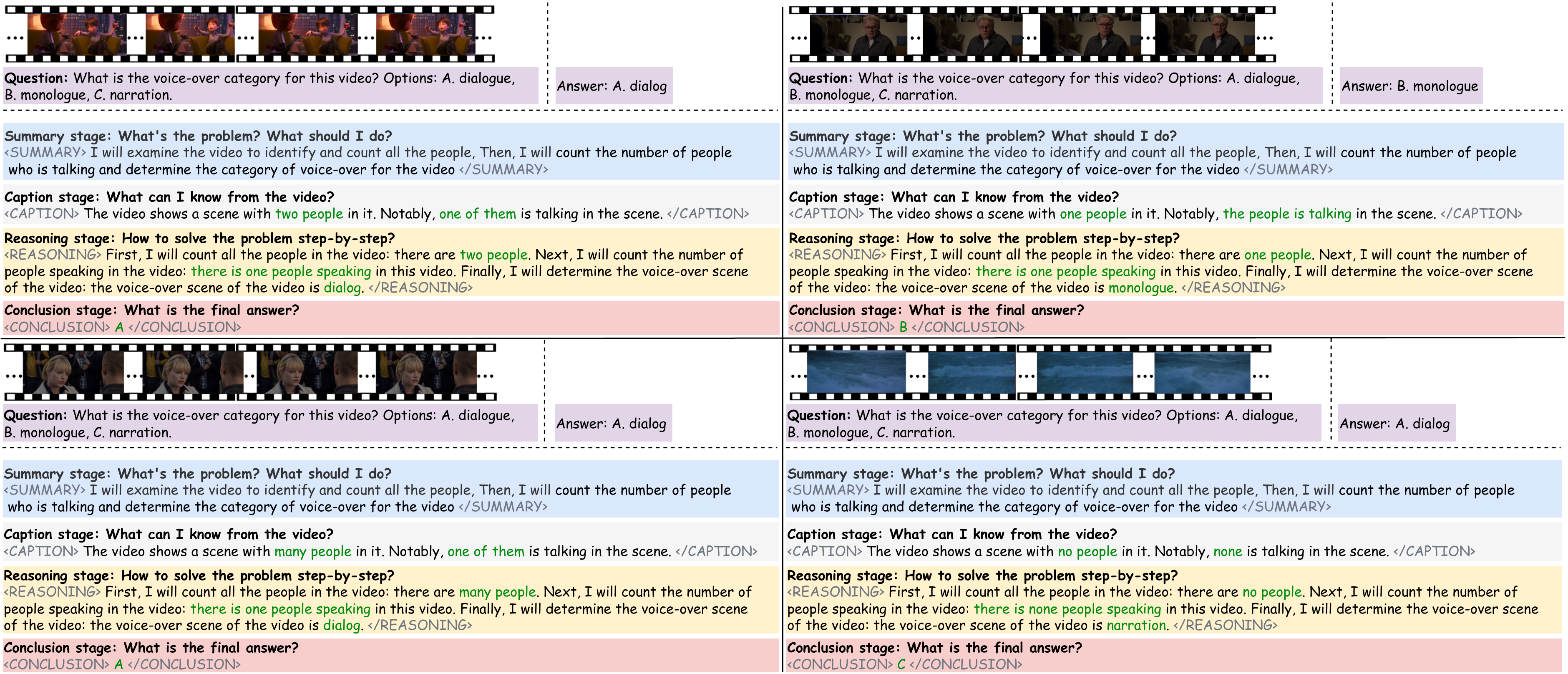}
    }
    \caption{The reasoning stages of movie scene type CoT annotations.}
    \label{fig:dataset_cot}
\end{figure*}

\begin{figure}[t!]
    \centering
    \resizebox{\linewidth}{!}{
    \includegraphics[]{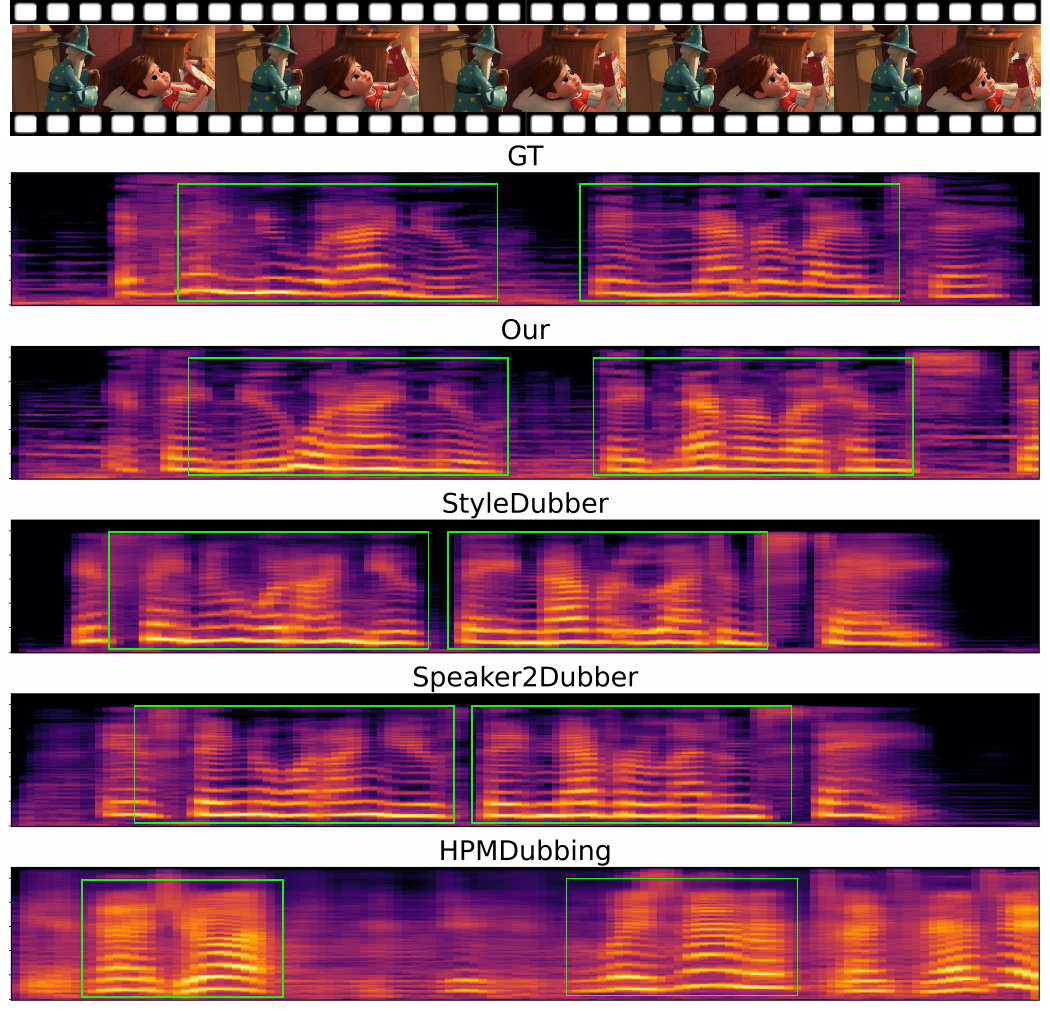}
    }
    \caption{Visualization of speech samples generated by state-of-the-art models and our. The green rectangles highlight key regions that have significant differences in overall expressiveness.}
    \label{fig:mel_visualization}
\end{figure}

\begin{table*}[t!]
\centering
\caption{
\textbf{Objective evaluation of the initial reasoning setting.}
% Reasoning Stage in Movie Scene Type Classification and Speech Generation.
For speech generation setting, we use the target speaker's speech as voice prompt if the predict scene type is correct and use random speaker's speech as voice prompt if the predict scene type is not correct.}
\resizebox{\linewidth}{!}{
    \begin{tabular}{ccccccccccc}
    \toprule
    \multicolumn{2}{c}{\multirow{2.5}{*}{Models Name}} & \multicolumn{5}{c}{Scores on dialogue(A), monologue(B) and narration(c)} & \multicolumn{4}{c}{Speech Generation} \\ \cmidrule(r){3-7} \cmidrule(r){8-11}
    \multicolumn{2}{c}{} & Ave.Acc(\%) $\uparrow$ & Ave.Recall(\%) $\uparrow$ & A.Recall(\%) $\uparrow$ & B.Recall(\%) $\uparrow$ & C.Recall(\%) $\uparrow$ & SPK-SIM(\%) $\uparrow$ & WER(\%) $\downarrow$ & MCD $\downarrow$ & MCD-SL $\downarrow$ \\ \midrule
    \multicolumn{11}{c}{MLLMs based} \\ \midrule
    \multirow{2}{*}{Qwen \cite{qwen2025qwen25technicalreport}} & MMLM-1B \cite{InternVL2_5-1B} & 84.09 & 82.97 & 86.50 & 68.40 & 94.00 & 83.17 & 23.60 & 8.59 & 8.60 \\
     & MMLM-4B \cite{InternVL2_5-4B} & 81.73 & 80.98 & 83.33 & \textbf{75.20} & 84.40 & 83.34 & 23.41 & 8.53 & 8.53 \\ \midrule
    \multirow{2}{*}{InternLM \cite{cai2024internlm2technicalreport}} & MMLM-2B \cite{InternVL2_5-2B} & 84.18 & 81.23 & \textbf{90.50} & 59.20 & 94.00 & 82.97 & 23.20 & 8.58 & 8.60 \\
     & MMLM-8B \cite{InternVL2_5-8B} & \textbf{86.00} & \textbf{85.84} & 86.33 & 73.20 & \textbf{98.00} & \textbf{83.42 \textcolor[HTML]{006400}{(+30.28\%)}} & \textbf{23.20 \textcolor[HTML]{006400}{(+55.70\%)}} & \textbf{8.54 \textcolor[HTML]{006400}{(+0.93\%)}} & \textbf{8.54 \textcolor[HTML]{006400}{(+3.94\%)}} \\ \midrule
    \multicolumn{11}{c}{Dubbing Models} \\ \midrule
     \multicolumn{2}{c}{HPMDubbing \cite{congLearningDubMovies2023}} & - & - & - & - & - & 61.06 & 199.40 & 8.82 & 11.88 \\
     \multicolumn{2}{c}{Speaker2Dub \cite{zhang2024from}} & - & - & - & - & - & 61.73 & 84.42 & 8.75 & 10.78 \\
     \multicolumn{2}{c}{StyleDubber \cite{cong2024styledubbermultiscalestylelearning}} & - & - & - & - & - & 64.03 & 52.69 & 8.62 & 8.89 \\ \bottomrule 
    \end{tabular}
}
\label{table:scene_and_speech_benchmark}
\end{table*}

\begin{table*}[]
\caption{\textbf{Ablation of objective evaluation under the initial reasoning setting on the proposed dataset.} For the speech generation setting, we use the target speaker's speech as a voice prompt if the predicted scene type is correct, and a random speaker's speech if the predicted scene type is incorrect. F\&O Reward refers to format reward and outcome reward.}
\resizebox{\linewidth}{!}{
    \begin{tabular}{ccccccccccc}
    \toprule
     &  & \multicolumn{5}{c}{Scores on dialogue(A), monologue(B) and narration(c)} & \multicolumn{3}{c}{Speech Generation} \\ \midrule
    Methods & Setting & Ave.Acc(\%) $\uparrow$ & Ave.Recall(\%) $\uparrow$ & A.Recall(\%) $\uparrow$ & B.Recall(\%) $\uparrow$ & C.Recall(\%) $\uparrow$ & SPK-SIM(\%) $\uparrow$ & WER(\%) $\downarrow$ & MCD $\downarrow$ & MCD-SL $\downarrow$ \\ \midrule
    \textbf{{\ul Base Model}} &  &  &  &  &  &  &  &  &  & \\
    InternVL2\_5-8B \cite{InternVL2_5-8B} & QA & 39.45\% & 44.24\% & 28.83\% & 5.20\% & 99.20\% & 81.73\% & 24.82\% & 8.91 & 8.91 \\ \midrule
    \textbf{{\ul Our Models}} &  &  &  &  &  &  &  &  &  & \\
    \multirow{2}{*}{SFT\cite{InternVL2_5-8B}} & QA & 82.27\% & 79.13\% & 89.00\% & 50.00\% & \textbf{98.40\%} & 82.89\% & 23.52\% & 8.66 & 8.66 \\
     & Reasoning & 85.09\% & 82.10\% & \textbf{91.50\%} & 56.80\% & 98.00\% & 83.34\% & 23.41\% & 8.55 & 8.55 \\
    MPO \cite{wang2024enhancingreasoningabilitymultimodal} & RL & 84.00\% & 81.36\% & 89.67\% & 56.40\% & 98.00\% & 83.02\% & 23.49\% & 8.59 & 8.60 \\
    MPO + F\&O Reward \cite{deepseekai2025deepseekr1incentivizingreasoningcapability} & RL & \textbf{86.00\% \textcolor[HTML]{006400}{(+4.53\%)}} & \textbf{85.84\% \textcolor[HTML]{006400}{(+8.50\%)}} & 86.33\% & \textbf{73.20\%} & 98.00\% & \textbf{83.42\% \textcolor[HTML]{006400}{(+0.64\%)}} & \textbf{23.20\% \textcolor[HTML]{006400}{(+1.36\%)}} & \textbf{8.54 \textcolor[HTML]{006400}{(+1.39\%)}} & \textbf{8.54 \textcolor[HTML]{006400}{(+1.39\%)}} \\ \bottomrule
    \end{tabular}
}
\label{table:ab_scene_benchmark}
\end{table*}

\begin{table*}[]
% \caption{\textbf{Objective evaluation of the Reasoning Stage} based on scores in dialogue (A), monologue (B), and narration (C) under two different level as explianed in \ref{sec:proposed_dataset}.
\caption{\textbf{Objective evaluation of the Reasoning Stage} based on scores in dialogue (A), monologue (B), and narration (C) under two different levels, as explained in \ref{sec:proposed_dataset}.
% . Setting 1 features scenes with both real people and anime characters, whereas Setting 2 includes only anime characters.
}
\centering
\resizebox{\linewidth}{!}{
    \begin{tabular}{cccccccccccc}
    \toprule
    \multicolumn{2}{c}{\multirow{2}{*}{Model Name}} & \multicolumn{5}{c}{Setting1} & \multicolumn{5}{c}{Setting2} \\ \cmidrule(r){3-7} \cmidrule(r){8-12}
    \multicolumn{2}{c}{} & Ave.Acc(\%) $\uparrow$ & Ave.Recall(\%) $\uparrow$ & A.Recall(\%) $\uparrow$ & B.Recall(\%) $\uparrow$ & C.Recall(\%) $\uparrow$ & Ave.Acc(\%) $\uparrow$ & Ave.Recall(\%) $\uparrow$ & A.Recall(\%) $\uparrow$ & B.Recall(\%) $\uparrow$ & C.Recall(\%) $\uparrow$ \\ \midrule
    \multicolumn{12}{c}{Closed Source} \\ \midrule
    \multicolumn{2}{c}{GPT-4o\cite{GPT-4o}} & - & - & - & - & - & \textbf{73.97} & \textbf{64.58} & \textbf{91.11} & 44.68 & 57.97 \\ \midrule
    \multicolumn{12}{c}{Open Source} \\ \midrule
    \multirow{2}{*}{Qwen \cite{qwen2025qwen25technicalreport}} & MMLM-1B \cite{InternVL2_5-1B} & 84.09 & 82.97 & 86.50 & 68.40 & 94.00 & 61.86 & 53.78 & 80.44 & 12.77 & 68.12 \\
     & MMLM-4B \cite{InternVL2_5-4B} & 81.73 & 80.98 & 83.33 & \textbf{75.20} & 84.40 & 62.63 & 55.51 & 79.56 & 15.96 & 71.01 \\ \midrule
    \multirow{2}{*}{InternLM \cite{cai2024internlm2technicalreport}} & MMLM-2B \cite{InternVL2_5-2B} & 84.18 & 81.23 & \textbf{90.50} & 59.20 & 94.00 & 53.09 & 51.37 & 61.33 & 15.96 & \textbf{76.81} \\
     & MMLM-8B \cite{InternVL2_5-8B} & \textbf{86.00} & \textbf{85.84} & 86.33 & 73.20 & \textbf{98.00} & 67.53 & 59.29 & 79.56 & \textbf{60.64} & 37.68 \\ \bottomrule
    \end{tabular}
}
\label{table:mmlm_base_benchmark}
\end{table*}

\begin{table*}[]
\caption{\textbf{Objective results on V2C-Animation and Grid benchmark.} For the Dub 1.0 setting, we use the ground truth speech as reference speech, for the Dub 2.0 setting, we use the non-ground truth speech from the same speaker within the dataset as the reference speech which is more aligned with practical usage in dubbing.}
\centering
\resizebox{\linewidth}{!}{
    \begin{tabular}{c|c|ccccccc|cccccc}
    \toprule
    benchmark & Setting & \multicolumn{7}{c|}{Dub 1.0} & \multicolumn{6}{c}{Dub2.0} \\ \midrule
    \multirow{8}{*}{V2C} & Methods & \multicolumn{1}{c|}{Visual} & SPK-SIM(\%) $\uparrow$ & WER(\%) $\downarrow$ & \multicolumn{2}{c}{EMO-SIM(\%) $\uparrow$} & MCD $\downarrow$ & MCD-SL$\downarrow$ & SPK-SIM(\%) $\uparrow$ & WER(\%) $\downarrow$ & \multicolumn{2}{c}{EMO-SIM(\%) $\uparrow$} & MCD $\downarrow$ & MCD-SL $\downarrow$ \\ \cmidrule{2-15} 
     & GT & \multicolumn{1}{c|}{-} & 100.00 & 17.38 & \multicolumn{2}{c}{100.00} & 0.00 & 0.00 & 100.00 & 17.38 & \multicolumn{2}{c}{100.00} & 0.00 & 0.00 \\ \cmidrule{2-15} 
     & F5-TTS \cite{chen2024f5ttsfairytalerfakesfluent} & \multicolumn{1}{c|}{\ding{55}} & 89.3 & 24.41 & \multicolumn{2}{c}{76.78} & 8.32 & 8.32 & 83.11 & 24.83 & \multicolumn{2}{c}{64.91} & 10.86 & 10.87 \\ \cmidrule{2-15} 
     & HPMDubbing \cite{congLearningDubMovies2023} & \multicolumn{1}{c|}{\ding{51}} & 73.64 & 151.02 & \multicolumn{2}{c}{39.85} & 8.59 & 8.32 & 73.01 & 150.83 & \multicolumn{2}{c}{34.69} & 9.11 & 12.15 \\
     & Speaker2Dub \cite{zhang2024from} & \multicolumn{1}{c|}{\ding{51}} & 82.15 & 31.23 & \multicolumn{2}{c}{65.92} & 10.68 & 11.21 & 79.53 & 31.28 & \multicolumn{2}{c}{59.71} & 11.16 & 11.70 \\
     & StyleDubber \cite{cong2024styledubbermultiscalestylelearning} & \multicolumn{1}{c|}{\ding{51}} & 82.48 & 27.36 & \multicolumn{2}{c}{66.24} & 10.06 & 10.52 & 79.81 & 26.48 & \multicolumn{2}{c}{59.08} & 10.56 & 11.05 \\ \cmidrule{2-15} 
     & Ours & \multicolumn{1}{c|}{\ding{51}} & \textbf{89.74} & \textbf{22.51} & \multicolumn{2}{c}{\textbf{78.88}} & \textbf{6.98} & \textbf{6.99} & \textbf{83.30} & \textbf{24.71} & \multicolumn{2}{c}{\textbf{64.93}} & \textbf{8.80} & \textbf{8.80} \\ \midrule
    \multirow{8}{*}{Grid} & Methods & \multicolumn{1}{c|}{Visual} & SPK-SIM(\%) $\uparrow$ & WER(\%) $\uparrow$ & LSE-C $\downarrow$ & LSE-D $\downarrow$ & MCD $\downarrow$ & MCD-SL $\downarrow$ & SPK-SIM(\%) $\uparrow$ & WER(\%) $\uparrow$ & LSE-C $\downarrow$ & LSE-D $\downarrow$ & MCD $\downarrow$ & MCD-SL $\downarrow$ \\ \cmidrule{2-15} 
     & GT & \multicolumn{1}{c|}{-} & 100.00 & 13.67 & 7.18 & 13.36 & 0.00 & 0.00 & 100.00 & 13.67 & 7.18 & 13.36 & 0.00 & 0.00 \\ \cmidrule{2-15} 
     & F5-TTS \cite{chen2024f5ttsfairytalerfakesfluent} & \multicolumn{1}{c|}{\ding{55}} & \textbf{96.51} & \textbf{11.94} & 5.51 & 14.70 & \textbf{4.23} & \textbf{4.24} & 94.45 & 16.75 & 5.10 & 14.71 & 4.89 & 4.90 \\ \cmidrule{2-15} 
     & HPMDubbing \cite{congLearningDubMovies2023} & \multicolumn{1}{c|}{\ding{51}} & 93.64 & 16.78 & \textbf{6.35} & 14.78 & 4.57 & 4.85 & 92.84 & 17.40 & \textbf{6.34} & 14.79 & 4.95 & 5.24 \\
     & Speaker2Dub \cite{zhang2024from} & \multicolumn{1}{c|}{\ding{51}} & 96.11 & 12.11 & 5.64 & 14.82 & 7.85 & 8.01 & 94.91 & 12.89 & 5.56 & 14.84 & 7.57 & 7.73 \\
     & StyleDubber \cite{cong2024styledubbermultiscalestylelearning} & \multicolumn{1}{c|}{\ding{51}} & 96.40 & 11.97 & 6.19 & 14.81 & 7.71 & 7.81 & \textbf{95.25} & \textbf{11.97} & 6.16 & 14.83 & 7.34 & 7.43 \\ \cmidrule{2-15} 
     & Ours & \multicolumn{1}{c|}{\ding{51}} & 95.73 & 14.71 & 4.87 & \textbf{14.63} & 4.48 & 4.49 & 94.71 & 16.08 & 4.46 & \textbf{14.63} & \textbf{4.73} & \textbf{4.74} \\ \bottomrule
    \end{tabular}
}
\label{table:benchmark}
\end{table*}

\begin{table}[]
\centering
\caption{\textbf{Subjective evaluation on V2C-Animation and GRID benchmarks.}}
\resizebox{\linewidth}{!}{
    \begin{tabular}{c|cc|cc}
    \toprule
    Dataset & \multicolumn{2}{c|}{V2C-Animation} & \multicolumn{2}{c}{GRID} \\ \midrule
    Methods & NMOS $\uparrow$ & SMOS $\uparrow$ & NMOS $\uparrow$ & SMOS $\uparrow$ \\ \midrule
    GT & 4.98±0.01 & - & 4.99±0.01 & - \\ \midrule
    F5-TTS \cite{chen2024f5ttsfairytalerfakesfluent} & 4.20±0.68 & 3.83±0.63 & \textbf{4.43±0.03} & \textbf{3.32±0.05}  \\ \midrule
    HPMDubbing \cite{congLearningDubMovies2023} & 1.04±0.01 & 1.02±0.01 & 3.50±0.10 & 2.77±0.12 \\
    Speaker2Dubber \cite{zhang2024from} & 2.93±0.21 & 2.58±0.19 & 4.04±0.07 & 3.00±0.10 \\
    StyleDubber \cite{cong2024styledubbermultiscalestylelearning} & 2.68±0.21 & 2.39±0.21 & 4.01±0.03 & 3.06±0.07 \\ \midrule
    Ours & \textbf{4.37±0.35} & \textbf{3.91±0.45} & 4.33±0.07 & 3.14±0.08 \\ \bottomrule
    \end{tabular}
}
\label{table:mos_benchmark}
\end{table}

\begin{table}[]
\centering
\caption{\textbf{Results on zero-shot test}, which use unseen speaker as reference speech.}
\resizebox{\linewidth}{!}{
    \begin{tabular}{c|cccccc}
    \toprule
    Setting & \multicolumn{5}{c}{Dubbing Setting 3.0} \\ \midrule
    Methods & LSE-C $\uparrow$ & LSE-D $\downarrow$ & SPK-SIM $\uparrow$ (\%) & WER(\%) $\downarrow$ & MOS $\uparrow$ \\ \midrule
    HPMDubbing \cite{congLearningDubMovies2023} & 1.72 & \textbf{11.74} & 68.14 & 126.85 & 1.29±0.60  \\
    Speaker2Dub \cite{zhang2024from} & 2.21 & 12.67 & 76.10 & 16.57 & 3.38±0.14 \\
    StyleDubber \cite{cong2024styledubbermultiscalestylelearning} & 2.15 & 12.76 & 78.30 & 19.07 & 3.30±0.15 \\ \midrule
    Ours & \textbf{2.21} & 12.59 & \textbf{83.55} & \textbf{15.49} & \textbf{4.12±0.16} \\ \midrule
    \end{tabular}
}
\label{table:zeroshot_test}
\end{table}

\begin{table}[ht]
\centering
\caption{\textbf{Results of ablation study on the proposed dataset with 2.0 setting.}}
\resizebox{\linewidth}{!}{
    \begin{tabular}{c|cccccc}
    \toprule
     & LSE-C $\uparrow$ & LSE-D $\downarrow$ & SPK-SIM (\%) $\uparrow$ & EMO-SIM (\%) $\uparrow$ & MCD $\downarrow$ & MCD-SL $\downarrow$  \\ \midrule
    w/o Clip & 1.99 & 12.73 & 82.63 & 63.24 & 8.85 & 8.86 \\
    w/o Dur & \textbf{2.06} & 12.81 & 82.71 & 64.32 & 8.82 & 8.83 \\ \midrule
    w/o Conclusion & 1.89 & 12.66 & 82.59 & 63.18 & 8.81 & 8.82 \\
    Proposed & 2.01 & \textbf{12.61} & \textbf{82.99} & \textbf{64.74} & \textbf{8.76} & \textbf{8.77} \\ \bottomrule
    \end{tabular}
}
\label{table:aba_res}
\end{table}

\section{Experiments}

\subsection{Datasets}

\noindent\textbf{Emilia~\cite{he2024emiliaextensivemultilingualdiverse}} is a comprehensive multilingual speech generation dataset containing a total of 101,654 hours of speech data across six languages. The English portion of this dataset, comprising approximately 46,800 hours, is utilized to train our foundational TTS model.

\noindent\textbf{V2C-Animation~\cite{chen2021v2cvisualvoicecloning}} is a specialized dataset designed for animated movie dubbing, consisting of 10,217 clips from 26 films with synchronized text, speech, and video. The dataset is partitioned into 60\% for training, 10\% for validation, and 30\% for testing.

\noindent\textbf{GRID} is a dubbing benchmark for multi-speaker dubbing~\cite{33222229005}. The whole dataset has 33 speakers, each with 1000 short English samples. All participants are recorded in studio with unified background. The number of train and test data are 32,670 and 3280, respectively.

\noindent\textbf{Chian-of-Thought Movie Dubbing Dataset.} We build a $7.2$ hour multimodal CoT movie dubbing dataset for generating high-quality and accurate movie dubbing. Based on CoT reasoning and CoT-like guidance \cite{xu2025llavacotletvisionlanguage}, we utilize a professional annotation team to label the following dataset. We develop a CoT reasoning framework to guide subsequent movie dubbing tasks, as illustrated in Figures \ref{fig:dataset} and \ref{fig:dataset_cot}. Specifically, a step-by-step instruction process with video input is designed to enable efficient and accurate movie scene type classification.
As shown in Figure 4, \texttt{<SUMMARY>}\texttt{</SUMMARY>} provides a high-level overview of the entire scene, while \texttt{<CAPTION>}\texttt{</CAPTION>} describes the characters in the video. During the \texttt{<REASONING>}\texttt{</REASONING>} stage, the reasoning process is divided into four steps:

\textbf{Step 1.} Count the numbers of people in the video.

\textbf{Step 2.} Distinguish whether the people in the video are talking or not.

\textbf{Step 3.} Distinguish whether the movie contains dialogue, narration, or monologue.

\textbf{Step 4.} Conclusion and give the answer.

And then \texttt{<CONCLUSION>}\texttt{</CONCLUSION>} stage give the final answer. Each stage is initiated at the model’s discretion, without external prompt engineering frameworks or additional prompting. Specifically, we provide the model with four pairs of special tags, these tags correspond to summarizing the response approach, describing relevant image content, conducting reasoning, and preparing a final answer, respectively. 
The proposed dataset consists of two difficulty levels: (1) Level-1, where people are talking in the videos, includes 7,276 video clips for training and 1,100 video clips for testing. (2) Level-2, where animals are talking in the videos, includes 3,486 video clips for training and 388 video clips for testing. Notably, due to OpenAI's fine-tuning policy, all Level-1 video clips have been filtered out, and 328 video clips remain for Level-2 training.
% The proposed dataset is consisted of two difficulty level, (1) level-1: there are people talking in the videos, 7276 video clips for training and 1100 video clips for testing. (2) level-2: there are animal talking in the videos, 3486 video clips for training and 388 video clips for testing. To be noted, due to opanai finetune policy, all of the video clips are filtered out for level-1 and 328 of the videos clips are remained for level-2.
\label{sec:proposed_dataset}

\subsection{Evaluation Metrics}
We evaluate using both objective and subjective metrics. To assess pronunciation accuracy, we use Word Error Rate (WER) with Whisper-V3 \cite{radford2022robustspeechrecognitionlargescale} as the ASR model. Timbre consistency is evaluated with speaker encoder cosine similarity (SPK-SIM) \cite{cong2024styledubbermultiscalestylelearning}. We also calculate mel cepstral distortion dynamic time warping (MCD) and speech length variance (MCD-SL) \cite{battenberg2020locationrelativeattentionmechanismsrobust} for spectral and length differences. Emotion similarity (EMO-SIM) is assessed using a speech emotion recognition model \cite{ye2023temporal}. For alignment with video, we use Lip Sync Error Distance (LSE-D) and Lip Sync Error Confidence (LSE-C) metrics on the Grid benchmark, based on the pre-trained SyncNet model \cite{chung2017out}. For subjective evaluation, we conduct human evaluations of the Mean Opinion Score (MOS) for naturalness (NMOS) and similarity (SMOS), rated on a 1-to-5 scale with 95\% confidence intervals. Following \cite{zhang2024from}, participants evaluate the dubbing quality of 30 randomly selected speech samples from each test set.

\subsection{Benchmark Results}
We compare our approach with a TTS model \cite{chen2024f5ttsfairytalerfakesfluent} and three recent video dubbing models. HPMDubbing \cite{congLearningDubMovies2023} introduces an emotional prosody adaptor that enables fine-grained alignment of the speaker's emotions. StyleDubber \cite{cong2024styledubbermultiscalestylelearning} , on the other hand, designs a multimodal phoneme-level style adaptor that generates stylized voice tones based on facial expressions. Speaker2Dubber \cite{zhang2024from} combines character emotions, phoneme prosody, and lip movements to ensure consistency in both prosody and duration throughout the dubbing process.

\noindent\textbf{Results on movie scene type reasoning and speech generation.}
% In order to better fit the real-world usage scenarios, we also conducted a test on the proposed dataset, which we call the towards fined-grained movie dubbing test. This setting just uses the silent video and related script as inputs (no reference speech) to measure the performance of the dubbing model for different dubbing types(Dialogue, Narration and Monologue) and fine-grained attributes(gender, age and emotion). The results are shown on Table \ref{table:initial_resoning}. Our method outperforms the SOTA dubbing methods (StyleDubber and Speaker2Dub) on SPK-SIM, EMO-SIM and WER. In details, the SPK-SIM improved from 64.03\% to 70.83\%, WER decreaded from 52.69\% to 27.68\%. Furthermore, the proposed method still maintains the competitive performance in speech-visual synchronization (see LSE-D), 11.88\%, which slightly lower than HPMDubbing(11.46\%).
As shown in Table \ref{table:scene_and_speech_benchmark}, the MMLM-8B achieves superior performance across all benchmarks in the classification of movie scene types. Our method outperforms the SOTA dubbing methods (StyleDubber and Speaker2Dub) on SPK-SIM, WER and MCD/MCD-SL. In detail, SPK-SIM improved from 64. 03\% to 70. 83\%, WER decreased from 52. 69\% to 27. 68\%. 
And, as shown in \ref{table:mmlm_base_benchmark}, the MMLM-8B maintains the competitive performance which slightly lower than GPT-4o \cite{GPT-4o}.
% Furthermore, the proposed method still maintains competitive performance in speech-visual synchronization (see LSE-D), 11.88\%, which is slightly lower than HPMDubbing(11.46\%).
These results demonstrate the effectiveness of our multimodal reasoning stages in enhancing multimodal movie dubbing performance.

\noindent\textbf{Results on V2C-Animation benchmark.}
As shown in Table \ref{table:mmlm_base_benchmark}, compared to the state-of-the-art models \cite{congLearningDubMovies2023,cong2024styledubbermultiscalestylelearning,zhang2024from}, our model achieves improvements across evaluation metrics in the same setting \cite{zhang2024from}. Our model achieves the best performance across all metrics. In detail, SPK-SIM increased from 79.81\% to 83.30\%, EMO-SIM improved from 59.71\% to 64.93\%, MCD decreased from 9.11 to 8.80, and WER decreased from 26. 48\% to 24. 71\%. It shows that our framework achieves performance improvement in pronunciation accuracy and consistency of speech duration.

%Since the V2C-Animation dataset is derived from real movie dubbing clips, its samples involve complex pronunciation and prosody variations, making it more challenging for the model to learn accurate pronunciation and duration consistency. Previous dubbing models fail to achieve accurate pronunciation, as reflected in the high WER values. However, our model significantly outperforms the others in pronunciation accuracy. In addition to pronunciation accuracy, our model also achieves the lowest MCD-DTW and MCD-DTW-SL. When compared to the F5-TTS model, our model slightly improves duration consistency with additional control.

\noindent\textbf{Results on GRID benchmark.}
As shown in Table \ref{table:benchmark}, our model achieves the best lip-sync performance on the GRID benchmark with the same evaluation of the state-of-the-art models \cite{zhang2024from}, which decreased from 14.79 to 14.63. And MCD decreased from 4.95 to 4.73.  Unlike V2C-Animation, samples in GRID are recorded in a studio environment, which does not involve exaggerated prosody variations or background noise. As a result, the WER of all comparison methods is generally better on the GRID compared to V2C-Animation. As shown in Table \ref{table:benchmark}, our model achieves the best lip-sync performance on the GRID benchmark. In addition, our method also achieves competitive results in WER, slightly lower than the best fine-tuned F5-TTS model, Speaker2Dub and StyleDubber. However, these models have WER results (11.94\%, 12.11\% and 11.97\%) that exceed the ground-truth WER result (13. 67\%), suggesting that the intelligibility has reached an acceptable range for humans.

\noindent\textbf{Results on Speaker Zero-shot test.}
As shown in Table \ref{table:zeroshot_test}, this setting uses the speech of unseen speakers as reference speech to measure the generalizability of the dubbing model \cite{zhang2024from}. Here, we use the speech from GRID as reference speech to measure V2C. We compare LSE-C/D, SPK-SIM, and WER, along with subjective evaluations in the same evaluation setting with the state-of-the-art models \cite{zhang2024from}. As shown in Table \ref{table:zeroshot_test}, our method outperforms StyleDubber and Speaker2Dub in both SPK-SIM and WER. In detail, the SIP-SIM improved from 78.30\% to 83.55\%, the WER decreased from 16.57\% to 15.49\%. Furthermore, the proposed method still maintains competitive performance in speech-visual synchronization (see LSE-C and LSE-D), slightly lower than HPMDubbing.
% we were surprised to find that This indicates that our model is more robust in maintaining clear pronunciation in unseen speaker scenes. 

% \noindent\textbf{Towards initial reasoning.}
% Results of "initial reasoning" test. In addition to the evaluation on proposed benchmarks, we also conduct a "initial reasoning" experiment to verify the robustness of our method. 
% In the "initial reasoning" experiment, we utilize scripts and movie clips from the proposed dataset, the input of the whole process is only the silent video and the corresponding subtitles to simulate the application of generating dubbed videos in real-world scenarios. The results was showed on Table~\ref{table:initial_resoning}.
% Due to the absence of corresponding ground truth in this test, we only calculate WER and SECS for objective evaluation to assess pronunciation quality and timbre consistency.

\subsection{Ablation Studies}
\noindent\textbf{Ablation Studies on Reasoning Stages.}
To compare the impact of SFT, MPO and MPO with F\&O rewards on improving multimodal reasoning ability, we used constructed CoT and QA pairs as training data to fine-tune InternVL2-8B. As shown in Table \ref{table:ab_scene_benchmark}, the results indicate that the model trained with MPO with F\&O rewards consistently outperforms that trained with Zeroshot, SFT and MPO. For example, the MPO (with F\&O rewards) trained model achieves an acc of 86.00\% on the movie scene reasoning benchmark, surpassing its SFT (QA) counterpart by 4.53\%. Furthermore, the MPO (with F\&O rewards) trained model also performs better on the recall rate of each category.

\noindent\textbf{Ablation studies on Speech Generation.}
% The ablation results in Table \ref{table:aba_res} indicate that each condition plays a role in overall performance. Removing the video clip control causes all metrics to drop significantly, showing its importance for speech-video alignment. Adding the video understanding conclusion control improves SPK-SIM and EMO-SIM. Finally, removing the duration predictor leads to the largest drop in LSE-D performance, highlighting that learning of duration-level consistency is essential to synchronizing speech and video. And with the MMLM conclusion conditions the SPK-SIM and EMO-SIM increased 0.4\% and 1.56\%, means the multimodal reasoning stages in enhancing multimodal movie dubbing performance.
The ablation results in Table \ref{table:aba_res} indicate that each condition contributes to overall performance. Removing the video clip control causes all metrics to drop significantly, highlighting its importance for speech-video alignment. Adding the video understanding conclusion control improves SPK-SIM and EMO-SIM. Furthermore, removing the duration predictor results in the largest drop in LSE-D performance, emphasizing that learning duration-level consistency is crucial for synchronizing speech and video. Additionally, with the MMLM conclusion conditions, SPK-SIM and EMO-SIM increase by 0.4\% and 1.56\%, respectively, demonstrating the effectiveness of multimodal reasoning stages in enhancing multimodal movie dubbing performance.

% The ablation results on proposed are presented in Table~\ref{table:aba_res_propoed}. It shows that both the two conditioning modules contribute to the overall performance, and each module has a different focus. After removing the duration predictor, the LSE severely up. This reflects that the duration predictor achieves better duration consistency by fusing video-level duration enhancement with explicit duration. In contrast, the mcd, emo-sim, speaker-sim is most affected by clip inputs, which indicates decoding mel-spectrograms by introducing frame-level video features is beneficial to overall speech performance.

\section{Conclusion}

In this paper, we propose a multi-stage, multimodal large language framework consisting of two-stage models and an accompanying multi-stage training strategy to improve the initial reasoning capabilities in movie dubbing. Additionally, we have created a corresponding movie dubbing dataset with CoT annotations. In the evaluation, the results show an improvement in performance compared to state-of-the-art methods across a variety of datasets.

{
    \small
    \bibliographystyle{ieeenat_fullname}
    \bibliography{main}
}

\end{document}